\documentclass{article}
\usepackage{cite}
\usepackage{url}
\usepackage{amsmath,amssymb,amsfonts}
\usepackage{algorithmic}
\usepackage{graphicx}
\usepackage{epsfig}
\usepackage{textcomp}
\usepackage{xcolor}
\usepackage{textpos}
\usepackage{hyperref}
\usepackage[hyphenbreaks]{breakurl}

\usepackage[normalem]{ulem}

\def\BibTeX{{\rm B\kern-.05em{\sc i\kern-.025em b}\kern-.08em
    T\kern-.1667em\lower.7ex\hbox{E}\kern-.125emX}}
\begin{document}

\title{A Neuromorphic Implementation of the DBSCAN Algorithm}

\author{Charles P. Rizzo \and James S. Plank}

\maketitle

\begin{center}
Department of Electrical Engineering and Computer Science \\
University of Tennessee \\
Knoxville, TN 37996 \\
\mbox{} \\
Corresponding authors: Charles P. Rizzo {\tt crizzo@utk.edu} or James S. Plank: {\tt jplank@utk.edu}
\end{center}

\begin{abstract}
DBSCAN is an algorithm that performs clustering in the presence of noise.
In this paper, we provide two constructions that allow DBSCAN to be implemented
neuromorphically, using spiking neural networks.  The first construction is
termed ``flat,'' resulting in large spiking neural networks that compute the
algorithm quickly, in five timesteps.  Moreover, the networks allow pipelining,
so that a new DBSCAN calculation may be performed every timestep.  The second
construction is termed ``systolic'', and generates much smaller networks, but
requires the inputs to be spiked in over several timesteps, column by column.
We provide precise specifications of the constructions and analyze them in
practical neuromorphic computing settings.  We also provide an open-source
implementation.
\end{abstract}


\section{Introduction}

The Density-based Spatial Clustering of Applications with Noise (DBSCAN) algorithm was first introduced in 1996, by Esker, Kriegel, Sander and 
Xu~\cite{eks:96:dbs}.  The algorithm takes as input a grid of bits, and performs
a clustering analysis on the grid.  Each one-bit is classified as a ``Core'' bit,
a ``Border'' bit or noise.  It has been an important algorithm in data mining,
winning the 2014 ``Test of Time'' award from SIGKDD, the ACM's special interest group
on knowledge discovery and data mining (\url{https://www.kdd.org/News/view/2014-sigkdd-test-of-time-award}).

Our interest in DBSCAN is its application to data from event-based cameras.  For example,
Nagaraj, Liyanagedera and Roy used DBSCAN in conjunction with a neuromorphic
speed filter to do bounding-box determination on event-based camera output~\cite{nlr:23:dotie}.  
Rizzo performed an expansion of this experiment and in doing so, started to develop
an implementation of
DBSCAN neuromorphically~\cite{r:24:eeb,rsp:24:sbf}.  This paper is the culmination of 
our explorations.

Neuromorphic computing uses spiking neurons and temporal synapses to perform processing.
It is attractive in embedded and edge-computing settings where processing with low power and SWaP
is a fundamental requirement.  Neuromorphic computing is especially attractive in conjunction
with event-based cameras,~\cite{lpd:06:128,i:19:346}
as events map natually to spikes, with no requirement to perform
encoding of input values into spikes, or decoding of spikes to output 
values~\cite{wgg:24:esn,d:21:atn,spb:19:nte}.  Recent research projects involving neuromorphic
computing and event-based cameras have involved estimation of optical 
flow~\cite{cpl:21:sso,crc:23:ofe}, gesture classification~\cite{rmh:23:drw}, 
eye-tracking~\cite{jwy:24:etb} and supervised image segmentation~\cite{hm:24:ens}.

In this paper, we specify two constructions of spiking neural networks to implement the
DBSCAN algorithm.  The first construction is termed ``flat,'' which results in large neural
networks that perform the algorithm very quickly.  With this construction, all of the inputs
to a DBSCAN problem are presented to the neural network at once, and the network calculates
the algorithm in five timesteps.  The algorithm pipelines perfectly, meaning new inputs may
be presented at every timestep. Therefore, there is no network refractory period between input applications.

As mentioned above, the flat algorithm results in large neural networks.  When processing
on a grid of~$R\times{C}$ pixels and evaluating radii of~$\epsilon$ units of the grid, the networks
contain $5RC$ neurons and roughly~$(3+2N)RC$ synapses, where~$N = (2\epsilon+1)^2$. $N$ is derived from the Chebyshev distance algorithm, which can be thought of as hop distance in all directions. In a grid, any point has $(2\epsilon+1)^2 - 1$ neighboring points reachable with a hop distance of $\epsilon$.
For example, a $260\times{346}$
grid, indicative of Inivation's DAVIS346 event-based camera~\cite{rsp:23:nde},
with $\epsilon = 4$ yields a network with nearly 450,000 neurons and over 14,000,000 
synapses.  Such a large network is impossible to deploy on today's neuromorphic hardware.

As an alternative, we also give a ``systolic'' construction.  With this construction, there
is a single input neuron for each row of the grid, and then the network consumes a column of
input at every timestep.  Processing input takes roughly~$C+2\epsilon$ timesteps, after which
new input may be applied to the network.  The networks are substantially smaller.
In the example of the preceding paragraph, the network is composed of 5,460 neurons and
46,700 synapses, which is feasible for implementation on current neuromorphic processors.

We have implemented these constructions in the TENNLab Neuromorphic Computing
Framework~\cite{psb:18:ten} on the RISP neuroprocessor~\cite{pzg:22:risp}.  We provide
our implementation as open-source, which produces networks that are not limited to the
TENNLab software/neuroprocessors,
but which may be employed on 
any leaky integrate-and-fire neuron system.  The networks
may be modified to work without leak, but are larger and require more timesteps.
We analyze some practical concerns
with these algorithms, including the ability
to partition the input grid to leverage smaller networks, and we also analyze some
practical concerns with communication and neuromorphic hardware.

Although we evaluate our networks on event-based data, we note that
these networks are not limited to event-based data.  These DBSCAN
networks can be applied to any spatially correlated data to which the
canonical DBSCAN algorithm may be applied.  Although it is out of
scope of this work, these networks may theoretically scale to higher
dimensional data as well. 

\section{DBSCAN Algorithm}

The DBSCAN algorithm operates on a~$R\times{C}$ grid of bits.  Since our application of
DBSCAN is on events from an event-based camera, we view the grid as being composed of
pixels that may or may not have events.  An event at a point or pixel in the grid
corresponds to the grid's bit being set to one.  The absence of an event corresponds to
the grid's bit being set to zero.  To ease the explanation of our algorithm, we present
it in terms of events and not bits, but the mapping from events to bits is a one-to-one
mapping.

The algorithm has two parameters: $\epsilon$ and $minPts$.  For each event (one bit) on the grid,
the algorithm classifies the event as one of three possible classifications: {\em Core}, {\em
Border} and {\em Noise}.  {\em Core} events are the center points of clusters and {\em Border}
events are found at the borders of clusters.  {\em Noise} events do not belong to any clusters,
and are typically discarded. 

The event classifications are determined as follows:

\begin{itemize}
\item Let $E_{r,c}$ be the event at row~$r$ and column~$c$ on the grid.
\item The neighborhood of $E_{r,c}$ is composed of all events~$E_{i,j}$ where
      $|i-r| \le \epsilon$ and $|j-c| \le \epsilon$.
\item Let~$N$ be the maximum size of a neighborhood of an event:
      $N = (2\epsilon+1)^2$.  $N$ is the maximum size, because neighborhoods
      are smaller at the borders of the grids, where, for example, an event on the top-left
      corner of the grid has no neighborhood above or left of it.
\item $minPts$ is a number between 1 and~$N$, and is a measure of the density of clustering.
\item An event is classified as {\em Core} if there are at least $minPts$ events
      in its neighborhood.
\item An event is classified as {\em Border} if it is not {\em Core}, but there is at least
      one {\em Core} event in its neighborhood.
\item An event is noise if it is neither {\em Core} nor {\em Border}.
\end{itemize}

\begin{figure}[ht]
\begin{center}
\begin{tabular}{ccc}
\epsfig{figure=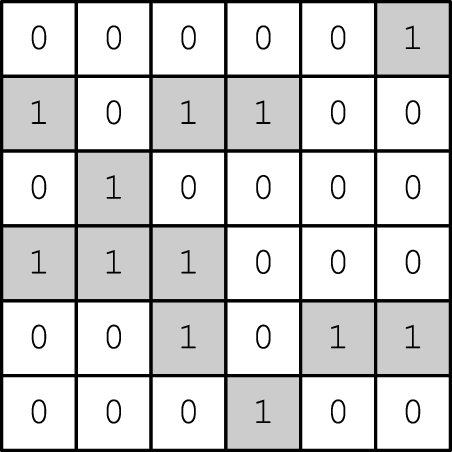,width=1.5in} &
\epsfig{figure=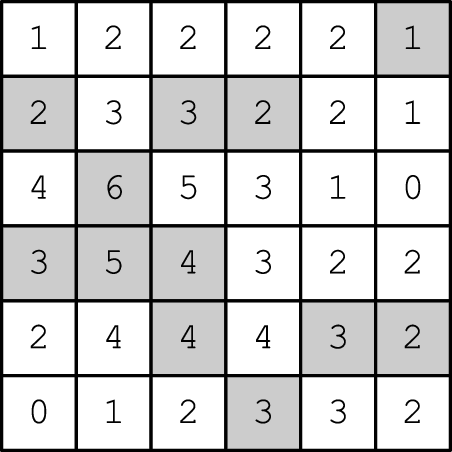,width=1.5in} &
\epsfig{figure=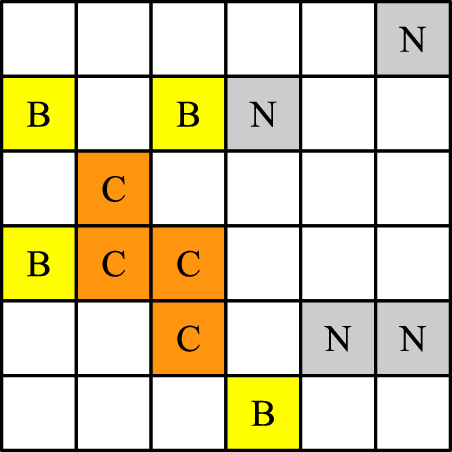,width=1.5in} \\
(a) & (b) & (c) \\
\end{tabular}
\caption{\label{fig:example} An example of DBSCAN with~$\epsilon = 1$ and $minPts = 4$: (a) A $6\times{6}$ input grid of events.
(b) The number of events in each neighborhood.
(c) The DBSCAN classification.  C = {\em Core}, B = {\em Border}, N = {\em Noise}.}
\end{center}
\end{figure}

We work through an example in Figure~\ref{fig:example}.  In part (a), we show an input
grid with 1's representing events, and 0's representing non-events.   In part (b),
we show the number of events in each grid point's neighborhood.  In part (c), we show the DBSCAN
classification of the events. 

\section{Properties of the Spiking Neural Network}

We employ a standard model of spiking neural networks.  Neurons are of the
leaky integrate-and-fire (LIF) variety.  Each neuron stores an activation potential
that is an integer between 0 and $(minPts-1)$.  Each neuron is configured with a
threshold.  If its activation potential meets or exceeds the threshold, then the neuron
spikes, and its potential is reset to zero.

Synapses connect pre-neurons to post-neurons, and are configured with two quantities: weight and delay.  Weights
are either one or negative one.  Delays are integers between 1 and 4.  When a synapse's 
pre-neuron spikes, the spike is transmitted by the synapse.  After the synapse's delay has
passed, the synapse's weight is added to its post-neuron's activation potential.

The neurons are processed in timesteps, or {\em integration cycles}.  Neurons only check their
activation potentials against their thresholds after all of their
incoming spikes have been processed for that timestep.

Neurons that do not fire may be configured to leak some or all of their activation 
potentials at the end of their integration cycles.  For this implementation of DBSCAN,
we configure the neurons so that they leak all of their potential to zero if they do
not fire.

Certain neurons may be designated as input neurons, and they receive input from the outside world.
For this implementation, the inputs are simple -- they force the neurons to spike at specific
timesteps.  Other neurons may be designated as output neurons.  The timesteps of their spikes
may be reported to the outside world.

This is a straightforward neuron model which may be implemented by most neuroprocessors that
support LIF neurons~\cite{spp:17:sur,rjp:19:twb}.  In this work, we use the RISP 
neuroprocessor~\cite{pzg:22:risp},
which is a very simple neuroprocessor from the TENNLab neuromorphic 
framework~\cite{psb:18:ten}.  Our open-source software produces neural networks that may
be translated easily to any neuroprocessor system with LIF neurons.

\section{The flat algorithm}

In the flat algorithm, there are five collections of neurons.  Each collection
contains~$RC$ neurons and is parameterized by~$r$ and~$c$, the row and column indices of the
neuron.  We assume that the grid on which the algorithm operates is composed of events~$E_{r,c}$.

We describe the five collections of neurons below:

\begin{itemize}
\item {\bf $I_{r,c}$ -- Inputs:}  These are neurons that receive inputs.  For each event in
the grid at row~$r$ and column~$c$, a spike is applied to neuron~$I_{r,c}$ at timestep zero that
forces~$I_{r,c}$ to fire.
\item {\bf $C_{r,c}$ -- Counts:}  These are neurons that count the number of 
events in the neighborhood of~$E_{r,c}$, not including the event~$E_{r,c}$.  Their
thresholds equal~$minPts-1$.  If the count
of $C_{r,c}$ is greater than or equal to~$(minPts-1)$, then~$C_{r,c}$ fires.
\item {\bf $Core_{r,c}$:}  These are neurons that fire when~$E_{r,c}$ is a {\em Core} event.
These are output neurons, and their thresholds are two.
\item {\bf $B_{r,c}$ -- Border-Cores:} These are neurons that count the number of {\em Core}
events in the neighborhood of~$E_{r,c}$, not including {\em Core} event~$E_{r,c}$. Their thresholds are one.  These neurons fire if there is a core event that
is not~$E_{r,c}$ but is in the neighborhood of~$E_{r,c}$.  
\item {\bf $Border_{r,c}$:}  These are neurons that fire when~$E_{r,c}$ is a {\em Border} event.
These are output neurons, and their thresholds are two.
\end{itemize}

\begin{figure}[ht]
\begin{center}
\begin{tabular}{ccccc}
\epsfig{figure=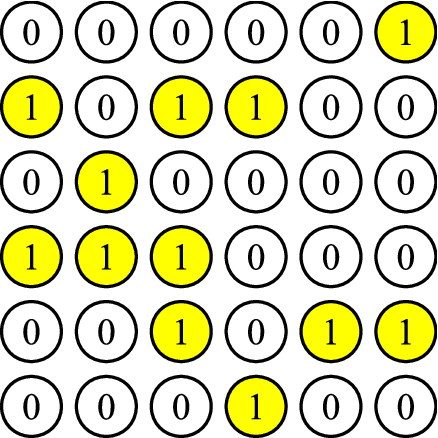,width=1.1in} &
\epsfig{figure=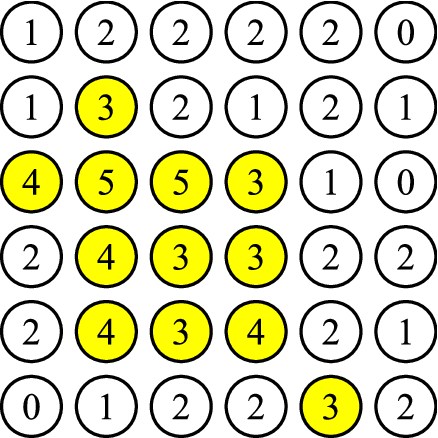,width=1.1in} &
\epsfig{figure=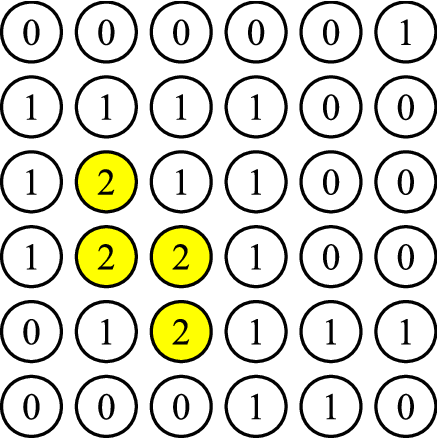,width=1.1in} &
\epsfig{figure=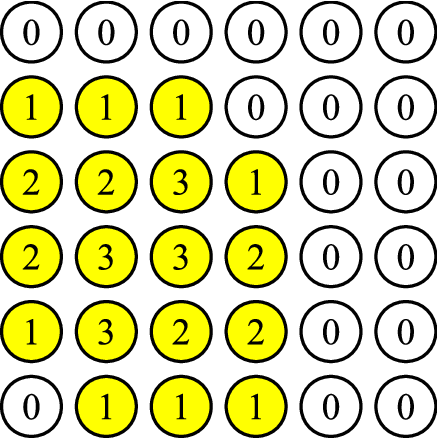,width=1.1in} &
\epsfig{figure=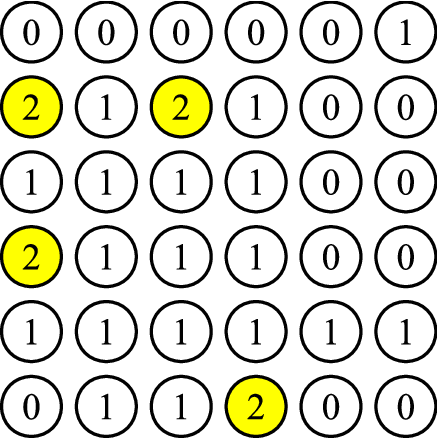,width=1.1in} \\
$I_{r,c}$ & $C_{r,c}$ & $Core_{r,c}$ & $B_{r,c}$ & $Border_{r,c}$ \\
Timestep 0 & Timestep 1 & Timestep 2 & Timestep 3 & Timestep 4 \\
\end{tabular}
\caption{\label{fig:neurons} The five collections of neurons for the example in Figure~\ref{fig:example}. The activation potential is shown at the relevant timestep, right before the neurons in yellow
spike.  Neurons that do not spike have their charge leak to zero before the next timestep.}
\end{center}
\end{figure}

To help illustrate these collections, in Figure~\ref{fig:neurons}, 
we show the neurons for the example above in Figure~\ref{fig:example}.
In the figure, we also show their activation potentials at the time during when they
fire.  If a neuron does not fire, then its activation potential leaks to zero at the
end of the timestep.

We now define the synapses so that the all of the neurons above spike correctly.

\begin{itemize}
\item There is a synapse from~$I_{i,j}$ to~$C_{r,c}$ if event~$E_{i,j}$ is in event~$E_{r,c}$'s
neighborhood, and $(i,j) \ne (r,c)$.  The synapse has weight 1 and delay 1.  These synapses 
cause neuron~$C_{r,c}$ to fire at timestep 1 whenever the number of events in~$E_{r,c}$'s
neighborhood (not including~$E_{r,c}$) is at least~$(minPts)-1$.
\item There is a synapse from~$I_{r,c}$ to~$Core_{r,c}$ with weight 1 and delay 2.
\item There is a synapse from~$C_{r,c}$ to~$Core_{r,c}$ with weight 1 and delay 1.  
Since $Core_{r,c}$ has a threshold of 2, it will only fire when it receives spikes from both
$I_{r,c}$ and~$C_{r,c}$.  That means it is a {\em Core} event.  It fires at timestep 2.
\item There is a synapse from~$Core_{i,j}$ to~$B_{r,c}$ if event~$E_{i,j}$ is in event~$E_{r,c}$'s
neighborhood, and $(i,j) \ne (r,c)$.  The synapse has weight 1 and delay 1.  These synapses 
cause neuron~$B_{r,c}$ to fire at timestep 3 whenever there is a core event, that is
not~$E_{r,c}$, in its neighborhood.
\item There is a synapse from~$I_{r,c}$ to~$Border_{r,c}$ with weight 1 and delay 4.
\item There is a synapse from~$Core_{r,c}$ to~$Border_{r,c}$ with weight -1 and delay 2.
\item There is a synapse from~$B_{r,c}$ to~$Border_{r,c}$ with weight 1 and delay 1.
Since $Border_{r,c}$ has a threshold of 2, it will only fire when it receives spikes from both
$I_{r,c}$ and~$B_{r,c}$, and no spike from~$Core_{r,c}$.  That means it is a {\em Border} event.  It fires at timestep 4.
\end{itemize}

In this construction, there are~$5RC$ neurons.  The number of synapses does not lend itself to an
easy calculation because neurons on the border have fewer incoming/outgoing synapses than those
in the middle.  Instead, in Table~\ref{tab:flatns}, 
we provide a rough estimate, which is on the high side, but should be
close enough for planning purposes.  To simplify these equations, we set~$N$, which is the number
of events in a neighborhood, to be: $(2\epsilon+1)^2$.

\begin{table}[ht]
\begin{center}
\begin{tabular}{|l|c|}
\hline
Synapses from~$I_{i,j}$ to~$C_{r,c}$ & $RC(N-1)$ \\
Synapses from~$I_{r,c}$ to~$Core_{r,c}$ & $RC$ \\
Synapses from~$C_{r,c}$ to~$Core_{r,c}$ & $RC$ \\
Synapses from~$Core_{i,j}$ to~$B_{r,c}$ & $RC(N-1)$ \\
Synapses from~$I_{r,c}$ to~$Border_{r,c}$ & $RC$ \\
Synapses from~$Core_{r,c}$ to~$Border_{r,c}$ & $RC$ \\
Synapses from~$B_{r,c}$ to~$Border_{r,c}$ & $RC$ \\
\hline
Total & $(3+2N)RC$ \\
\hline
\end{tabular}
\caption{\label{tab:flatns} Number of synapses in the flat DBSCAN network.}
\end{center}
\end{table}

\section{The systolic algorithm}

With the systolic algorithm, we reduce the sizes of the neuron collections, and instead of
spiking in all inputs at once, we spike~$R$ inputs at a time, over~$C$ timesteps.
The result is a much smaller network, and a reduced communication burden for inputs
and outputs.  This comes at the expense of speed.  The name ``systolic'' comes from
systolic VLSI arrays from the 1980's, which process data in a similar, 
stream-like fashion~\cite{hl:80:sav,bk:84:sva}.

The collections of neurons have the same names and functions as in the flat specification.
However, there are fewer of them, and instead of performing one large calculation every
timestep, they perform a column's worth of calculations every timestep.  In the description
below, we specify the neuron collections, and the synapses that go to each collection.
We illustrate with the same example as above.

\subsection{$I_{r,e}$ -- Inputs} 
\label{sec:i}

There are~$R(2\epsilon+1)$ of these neurons, labeled as follows:

\begin{eqnarray*}
r & | & 0 \le r < R \\
e & | & -\epsilon \le e < \epsilon \\
\end{eqnarray*}

Their thresholds are one.  The only neurons that are input neurons are~$I_{r,e}$
where~$e=\epsilon$.  Event~$E_{r,c}$ is applied so that neuron~$I_{r,\epsilon}$ spikes
at timestep~$c$.

For the other neurons,~$I_{r,e}$ where~$e \ne \epsilon$, there is a synapse from~$I_{r,e+1}$
with a weight of one and a delay of one.  Therefore, each event~$E_{r,c}$ causes~$I_{r,e}$
to spike at time~$c+\epsilon-e$.  In particular,~$E_{r,c}$ causes neuron~$I_{r,0}$ to
spike at time~$c+\epsilon$.

\begin{figure}[ht]
\begin{center}
\begin{tabular}{ccc}
\epsfig{figure=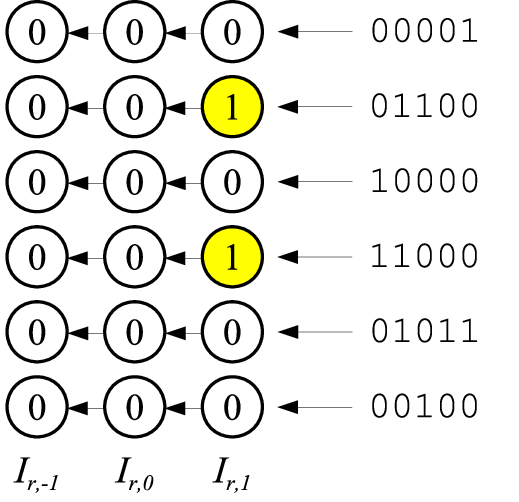,height=1.5in} &
\epsfig{figure=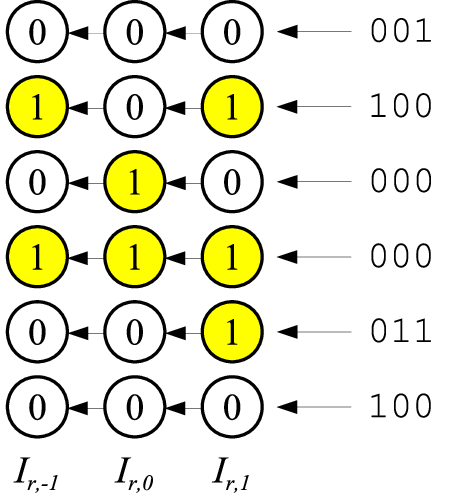,height=1.5in} &
\epsfig{figure=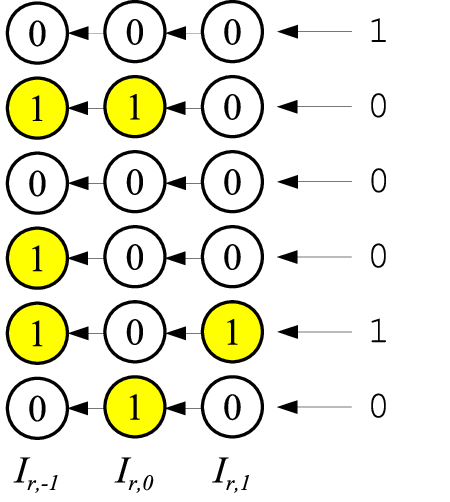,height=1.5in} \\
Timestep $T$=0 & Timestep $T$=2 & Timestep $T$=4 \\
\end{tabular}
\caption{\label{fig:syst_i} The $I$ neurons in the systolic algorithm for the example.  Instead of all inputs being applied at timestep 0, there is one input neuron per row ($I_{r,1}$), and the inputs are applied one column at a time.}
\end{center}
\end{figure}

In Figure~\ref{fig:syst_i}, we show the $I$ neurons at three different times in the
DBSCAN calculation of the example from the last section.  On the left, we show the
state of the network at the end of timestep $T=0$, where one column of inputs have been
applied to the network.  In the middle, we
show the state at the end of timestep $T=2$, with three columns of inputs applied, and on the
right, we show the state at the end of timestep $T=4$, with five columns of inputs applied. It can be helpful to think of the systolic network as ``scanning'' or convolving, with a stride of 1, column-wise across the grid.

The important feature to note here is that at the end of timestep~$(c+\epsilon)$,
column~$c$ is in the
middle column of the~$I$ neurons, at~$I_{r,0}$.
Therefore, the entire neighborhoods of the events
in column~$c$ are stored in the~$I$ neurons at timestep~$(c+\epsilon)$.
We use this fact to calculate~$C_{r,c}$ in the~$C$ neurons at timestep~$(c+\epsilon+1)$.
We describe this in the next section.

To illustrate, the middle of Figure~\ref{fig:syst_i} shows timestep 2, which equals
$(1+\epsilon)$.  Therefore, in that timestep, the neurons spike for all of the events in the
neighborhoods of~$E_{0,1}$ through~$E_{5,1}$.  Similarly, the right
side of Figure~\ref{fig:syst_i}, shows timestep 4, which equals
$(3+\epsilon)$.  Therefore, in that timestep, the neurons spike for all of the events in the
neighborhoods of~$E_{0,3}$ through~$E_{5,3}$.  

To help in the remainder of this explanation, the following table summarizes the meanings
of spikes in the~$I_{r,e}$ neurons:

\begin{center}
\begin{tabular}{|l|c|c|}
\hline
Neuron & Timestep & Meaning of Fire\\
\hline
$I_{r,\epsilon}$ & $c$ & There is an event~$E_{r,c}$. \\
$I_{r,0}$ & $c+\epsilon$ & There is an event~$E_{r,c}$. \\
$I_{r,-\epsilon}$ & $c+2\epsilon$ & There is an event~$E_{r,c}$. \\
$I_{r,e}$ & $c+\epsilon$ & This event occurs in neighborhoods of~$E_{i,c}$ 
                          for $|i-r| \le \epsilon$. \\
\hline
\end{tabular}
\end{center}

\subsection {\bf $C_{r}$ -- Counts}
\label{sec:c}

There are only~$R$ of these neurons, and their thresholds are~$(minPts-1)$.  
There is a synapse from~$I_{i,j}$ to~$C_{r}$ for
      $|i-r| \le \epsilon$ and $|j-c| \le \epsilon$, and~$(i,j) \ne (r,c)$.
With these synapses, the delays and weights are both one.

Therefore, at timestep~$(c+\epsilon+1)$, the action potential of neuron~$C_r$ is the same
as~$C_{r,c}$ in timestep 1 of the flat algorithm.  The neuron spikes at this time if
there are at least~$minPts-1$ events in the neighborhood of~$E_{r,c}$, not including~$E_{r,c}$.

\begin{figure}[ht]
\begin{center}
\epsfig{figure=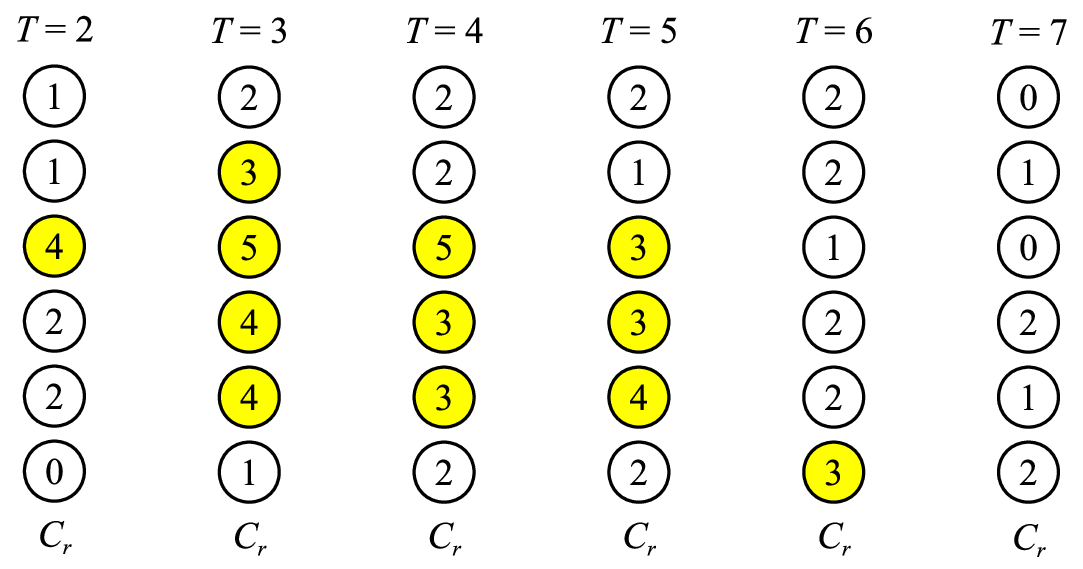,height=1.7in}
\caption{\label{fig:syst_c} The $C$ neurons at timesteps $T=2$ through $T=7$, corresponding to
columns $c=0$ through $c=5$ of the inputs.
At timestep~$c+\epsilon+1$, the neuron~$C_r$ spikes if there are
at least~$minPts-1$ events in the neighborhood of~$E_{r,c}$, not including~$E_{r,c}$.}
\end{center}
\end{figure}

In Figure~\ref{fig:syst_c}, we show the~$C$ neurons at their six relevant timesteps in the
example.  Note how the neurons match the~$C_{r,c}$ neurons in Figure~\ref{fig:neurons}.
Note also how the neurons at timestep~$T=3$ are calculated from the~$I$ neurons in the 
middle of Figure~\ref{fig:syst_i}, where~$T=2$.  Similarly, 
note how the neurons at timestep~$T=5$ are calculated from the~$I$ neurons in the 
right side of Figure~\ref{fig:syst_i}, where~$T=4$.

The following table summarizes the meaning of the spikes of the~$C$ neurons:

\begin{center}
\begin{tabular}{|l|c|c|}
\hline
Neuron & Timestep & Meaning of Fire\\
\hline
$C_{r}$ & $c+\epsilon+1$ & There are at least $minPts-1$ events in the neighborhood of~$E_{r,c}$, not including~$E_{r,c}$. \\
\hline
\end{tabular} 
\end{center}

\subsection{$Core_{r,c}$ -- {\em Core} events}
\label{sec:core}

There are $R(2\epsilon+1)$ {\em Core} neurons with the same labeling as the~$I$ neurons.
These are defined so that if~$E_{r,c}$ is a {\em Core} event, then~$Core_{r,\epsilon}$ fires at 
timestep~$c+\epsilon+2$.  Like the~$I$ neurons, the~$Core$ neurons are set up so that 
if~$Core_{r,e}$ spikes at timestep~$t$, then~$Core_{r,e-1}$ spikes
at timestep~$t+1$.  This propagates the {\em Core} events down the columns of the {\em Core}
neurons in the same manner that the input events are propagated down the columns of the~$I$
neurons.

Only the~$Core_{r,\epsilon}$ neurons are output neurons, with thresholds of two.  The other
$Core_{r,e}$ neurons, for~$e \ne \epsilon$, are hidden neurons with thresholds of one.

There are two synapses coming into the~$Core_{r,\epsilon}$ neurons:

\begin{enumerate}
\item A synapse from~$C_{r}$ with a weight of 1 and delay of 1.
\item A synapse from~$I_{r,0}$ with a weight of 1 and delay of 2.
\end{enumerate}

Thus, if~$Core_{r,\epsilon}$ spikes at timestep~$(c+\epsilon+2)$, it is because~$C_r$
spiked at time~$c+\epsilon+1$, and~$I_{r,0}$ spiked at time~$c+\epsilon$. From
Sections~\ref{sec:i} and~\ref{sec:c} above, that means that there is an event~$E_{r,c}$,
and there are at least~$minPts-1$ other events in the neighborhood of~$E_{r,c}$.
Therefore,~$E_{r,c}$ is a {\em Core} event.

For the other~$Core_{r,e}$ neurons, each has a synapse from~$Core_{r,e+1}$ with a weight
of 1 and a delay of 1.  Therefore, at timestep~$(c+2\epsilon+2)$, all of the core neurons
are in the neighborhoods of the~$E_{r,c}$ events.  This fact is used to define the {\em
Border} events.

\begin{figure}[ht]
\begin{center}
\begin{tabular}{ccc}
\epsfig{figure=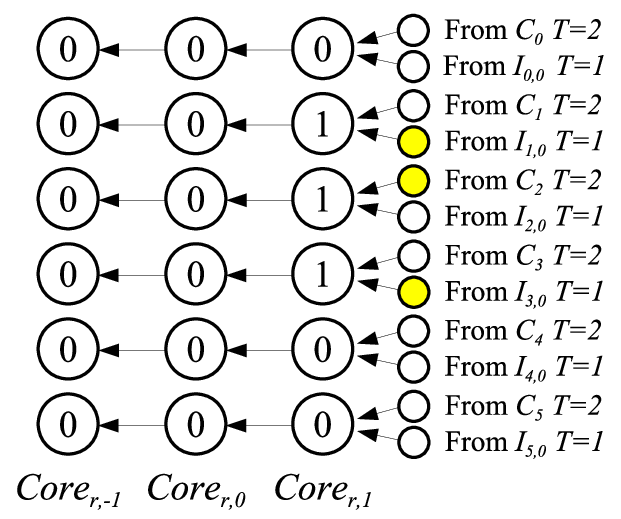,height=1.7in} &
\epsfig{figure=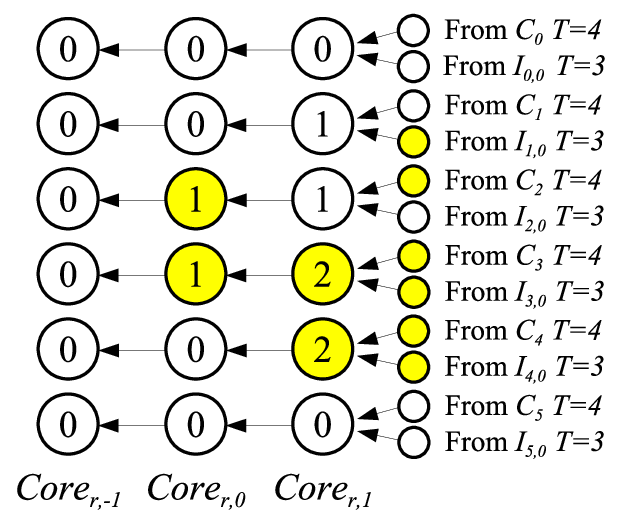,height=1.7in} &
\epsfig{figure=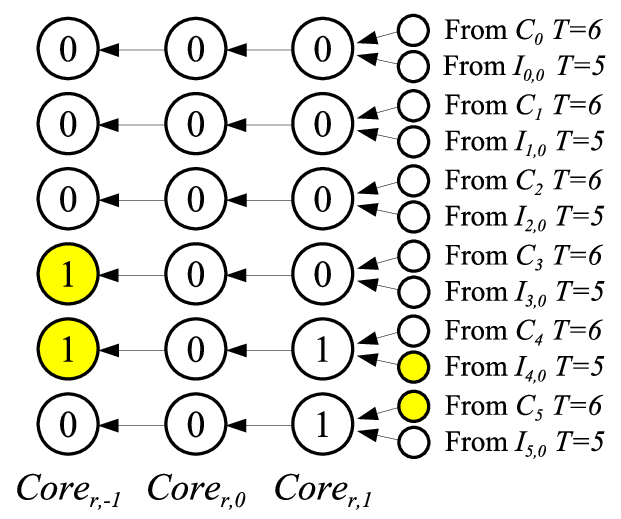,height=1.7in} \\
Timestep $T$=3 & Timestep $T$=5 & Timestep $T$=7 \\
\end{tabular}
\caption{\label{fig:syst_core} The $Core$ neurons in the systolic algorithm for the example.
At time~$c+\epsilon+2$, neuron~$Core_{r,\epsilon}$ spikes if~$E_{r,c}$ is a {\em Core}
event.}
\end{center}
\end{figure}

Figure~\ref{fig:syst_core} shows the {\em Core} neurons in three timesteps of our example.
The left side shows timestep~$T=3$.  Since~$3=(c+\epsilon+2)$, this corresponds to
the events in column~$c=0$.  
Since none of the~$Core_{r,\epsilon}$ neurons 
spike, there are no core events in column 0.  
The middle part of the figure shows
timestep~$T=5$, corresponding to~$c=2$.  Since neurons~$Core_{3,\epsilon}$ and
$Core_{4,\epsilon}$ spike at this timestep, we know that two events in column 2 -- ~$E_{3,2}$ and~$E_{4,2}$ -- are
core events.  Note how this matches column 2 in the $Core$ neurons of 
Figure~\ref{fig:neurons}.  The right part of the figure shows
timestep~$T=7$, corresponding to~$c=4$.  Since none of the~$Core_{r,\epsilon}$ neurons are 
spiking, there are no core events in column 4.

The {\em Core} neurons in the middle of Figure~\ref{fig:syst_core}
show the {\em Core} events in the neighborhoods of the events~$E_{r,1}$.  
That is because the timestep is $5 = (c+2\epsilon+2)$ and therefore~$c=1$.
This will be used to calculate the events~$E_{r,1}$ that are {\em Border} events.

The following table summarizes the meaning of the spikes of the~$Core$ neurons:

\begin{center}
\begin{tabular}{|l|c|c|}
\hline
Neuron & Timestep & Meaning of Fire\\ 
\hline 
$Core_{r,\epsilon}$ & $c+\epsilon+2$ & $E_{r,c}$ is a {\em Core} event. \\
$Core_{r,0}$ & $c+2\epsilon+2$ & $E_{r,c}$ is a {\em Core} event. \\
$Core_{r,e}$ & $c+2\epsilon+2$ & This core event occurs 
in neighborhoods of~$E_{i,c}$ for $|i-r| \le \epsilon$. \\
\hline
\end{tabular} 
\end{center}

\subsection{$B_{r}$ -- Border-Cores:}
\label{sec:b}  

Like the~$C$ neurons, there are only~$R$ of these neurons. Their thresholds
are one.  There is a synapse from~$Core_{i,j}$ to~$B_r$ for
$|i-r| \le \epsilon$ and $|j-c| \le \epsilon$, and~$(i,j) \ne (r,c)$.

The~$B_r$ neurons have the same function as the~$B_{r,c}$ neurons in the flat algorithm.
If there is any {\em Core} event in the neighborhood of~$E_{r,c}$, with the exception
of~$E_{r,c}$, then~$B_r$ fires
at time~$(c+2\epsilon+3)$.

\begin{figure}[ht]
\begin{center}
\epsfig{figure=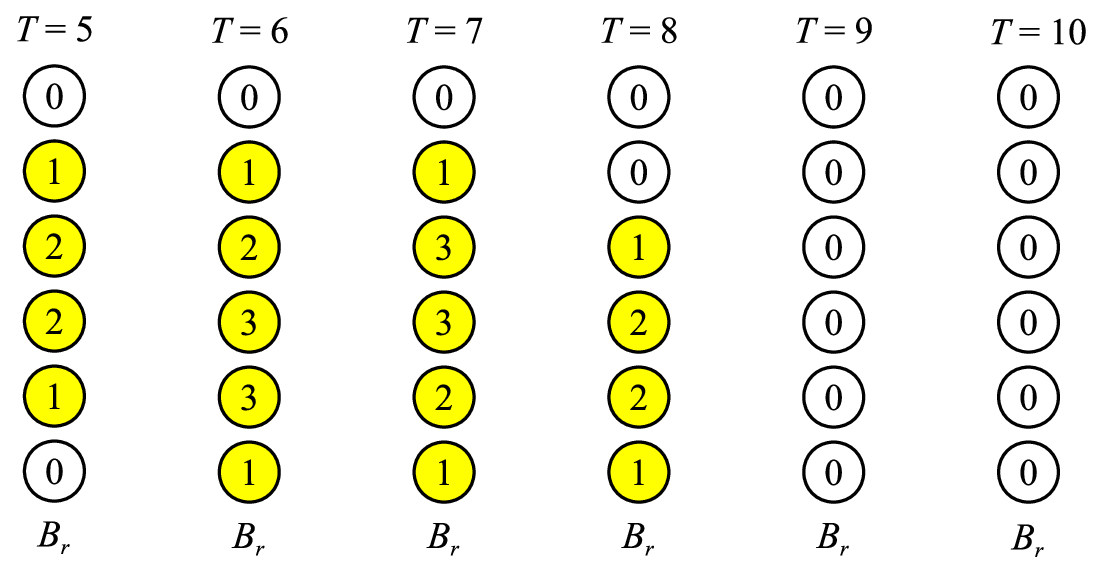,height=1.7in}
\caption{\label{fig:syst_b} The $B$ neurons at timesteps 5 to 10.  These timesteps
correspond to~$c$ going from 0 to 5 in~$c+2\epsilon+3$.}
\end{center}
\end{figure}

Figure~\ref{fig:syst_b} continues the example, showing the six relevant timesteps when
the~$B$ neurons fire.  
Note the similarity of~\ref{fig:syst_b} to the~$B$ neurons in Figure~\ref{fig:neurons}.

The following table summarizes the meaning of the spikes of the~$B$ neurons:

\begin{center}
\begin{tabular}{|l|c|c|}
\hline
Neuron & Timestep & Meaning of Fire\\
\hline 
$B_{r}$ & $c+2\epsilon+3$ & There is a {\em Core} event 
in neighborhood of~$E_{r,c}$, that is not~$E_{r,c}$. \\
\hline 
\end{tabular} 
\end{center} 

\subsection{$Border_{r}$:} 
\label{sec:border}

There are~$R$ of these neurons, which
are output neurons with thresholds of two.  There are three synapses coming into each~$Border_r$:

\begin{enumerate}
\item A synapse from~$B_r$ with a weight of 1 and a delay of 1.
\item A synapse from~$Core_{r,0}$ with a weight of -1 and a delay of 2.
\item A synapse from~$I_{r,-\epsilon}$ with a weight of 1 and a delay of~$4$.
\end{enumerate}

\noindent Therefore, if~$Border_{r}$ spikes at time~$(c+2\epsilon+4)$, it does so because:

\begin{itemize}
\item $I_{r,-\epsilon}$ spiked at time~$(c+2\epsilon)$.  From the table at the end of
      section~\ref{sec:i}, That means there is an event~$E_{r,c}$.
\item $B_{r}$ spiked at time~$(c+2\epsilon+3)$.  From the table at the end of
      section~\ref{sec:b}, that means there is a {\em Core} event in the
      neighborhood of~$E_{r,c}$.
\item $Core_{r,0}$ did not spike at time~$(c+2\epsilon+c)$. From the table at the end of
      section~\ref{sec:core}, that means that~$E_{r,c}$ is not a {\em Core event}.
\end{itemize}

\noindent Therefore,~$E_{r,c}$ is a {\em Border} event.

\begin{figure}[ht]
\begin{center}
\epsfig{figure=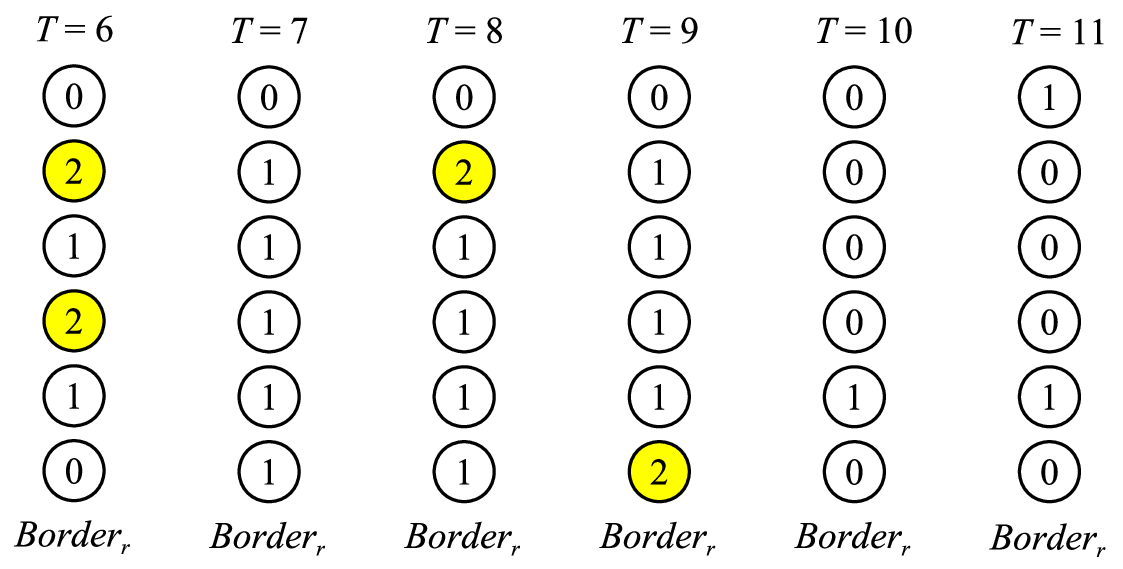,height=1.7in}
\caption{\label{fig:syst_border} The $Border$ neurons at timesteps 6 to 11.  These timesteps
correspond to~$c$ going from 0 to 5 in~$c+2\epsilon+4$.}
\end{center}
\end{figure}

Figure~\ref{fig:syst_border} shows the~$Border$ neurons for the example from timesteps 6 to 11.
Thus, they show the border events in the example.  Note how they match the~$Border$ neurons
in Figure~\ref{fig:neurons}.

The following table summarizes the firing of the {\em Border} neurons:

\begin{center}
\begin{tabular}{|l|c|c|}
\hline
Neuron & Timestep & Meaning of Fire\\
\hline
$Border_{r}$ & $c+2\epsilon+4$ & $E_{r,c}$ is a {\em Border} event. \\
\hline
\end{tabular}
\end{center}

\subsection{Systolic Network Reuse}

To reuse a systolic network for a subsequent DBSCAN calculation, one must apply no
input spikes for~$2\epsilon$ timesteps.  Therefore, the same network may perform DBSCAN
calculations every~$C+2\epsilon$ timesteps.

\subsection{Analysis}

The~$I$ and~$Core$ neurons number~$R*(2\epsilon+1)$ each, while the~$C$,
$B$ and $Border$ neurons number~$R$ each.  That is a total of $R(4\epsilon+5)$.
Table~\ref{tab:systolicns} shows the rough calculation for the number of
synapses.  As with the flat networks, we don't provide an exact calculation, but
instead assume that every neuron has the maximum number of outgoing synapses.

\begin{table}[ht]
\begin{center}
\begin{tabular}{|l|c|}
\hline
Synapses from~$I_{r,e}$ to~$I_{r,e-1}$ & $2R\epsilon$ \\
Synapses from~$I_{i,e}$ to~$C_{r}$ & $R(N-1)$ \\
Synapses from~$I_{r,0}$ to~$Core_{r,e}$ & $R$ \\
Synapses from~$I_{r,-\epsilon}$ to~$Border_{r}$ & $R$ \\
Synapses from~$C_{r}$ to~$Core_{r,e}$ & $R$ \\
Synapses from~$Core_{r,e}$ to~$Core_{r,e-1}$ & $2R\epsilon$ \\
Synapses from~$Core_{i,e}$ to~$B_{r}$ & $R(N-1)$ \\
Synapses from~$Core_{r,0}$ to~$Border_{r}$ & $R$ \\
Synapses from~$B_{r}$ to~$Border_{r}$ & $R$ \\
\hline
Total & $R(2N+4\epsilon+3) = R(8\epsilon^2 + 12\epsilon + 5)$ \\
\hline
\end{tabular}
\caption{\label{tab:systolicns} Number of synapses in the systolic DBSCAN network.}
\end{center}
\end{table}

\section{Partial DBSCAN}
\label{sec:partial}

There may be times where a single DBSCAN network, either flat or systolic, is too big
to fit on a neuroprocessor.  In that case, one may perform DBSCAN multiple
times, on smaller regions of the input space.   Doing this is not as simple as, for example,
partitioning a 100x100 grid into four 50x50 grids and performing DBSCAN on them individually.
This is because an event on the border of one of these subgrids may need information about
events in other subgrids.

We first consider the flat algorithm.  Suppose our input grid is~$R \times C$, and that
we wish to perform DBSCAN on a smaller, $I_R \times I_C$ region.  To do this,
we need to perform DBSCAN on a $(4\epsilon+I_R)\times(4\epsilon+I_C)$ subgrid, that contains
the~$I_R \times I_C$ region in its center.  

\begin{figure}[ht]
\begin{center}
\begin{tabular}{cc}
\epsfig{figure=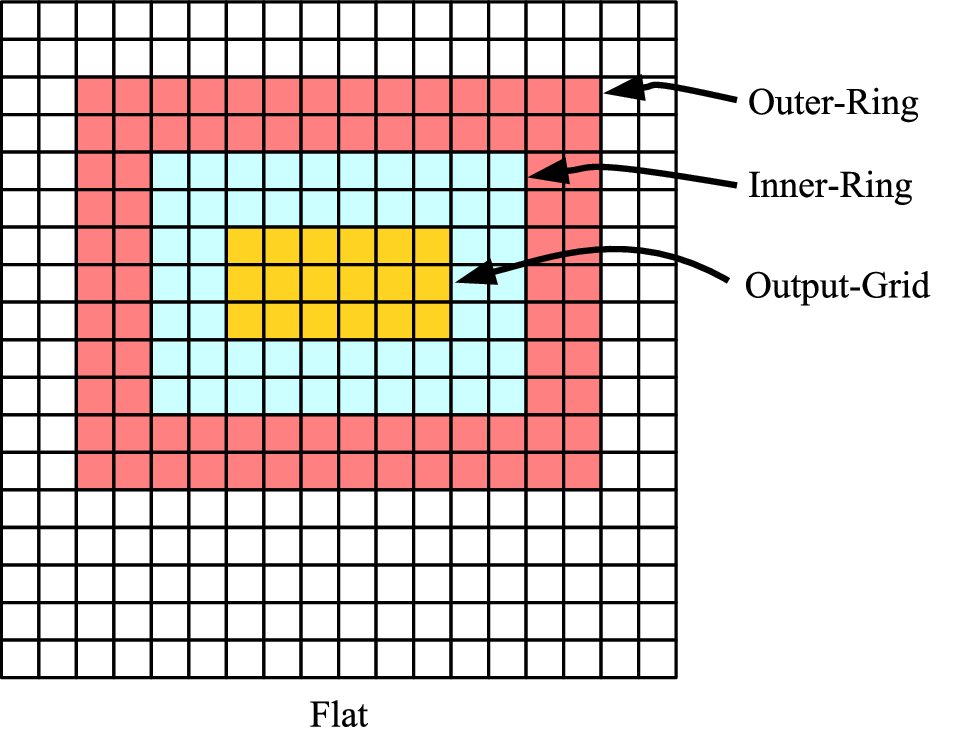,height=2.2in} &
\epsfig{figure=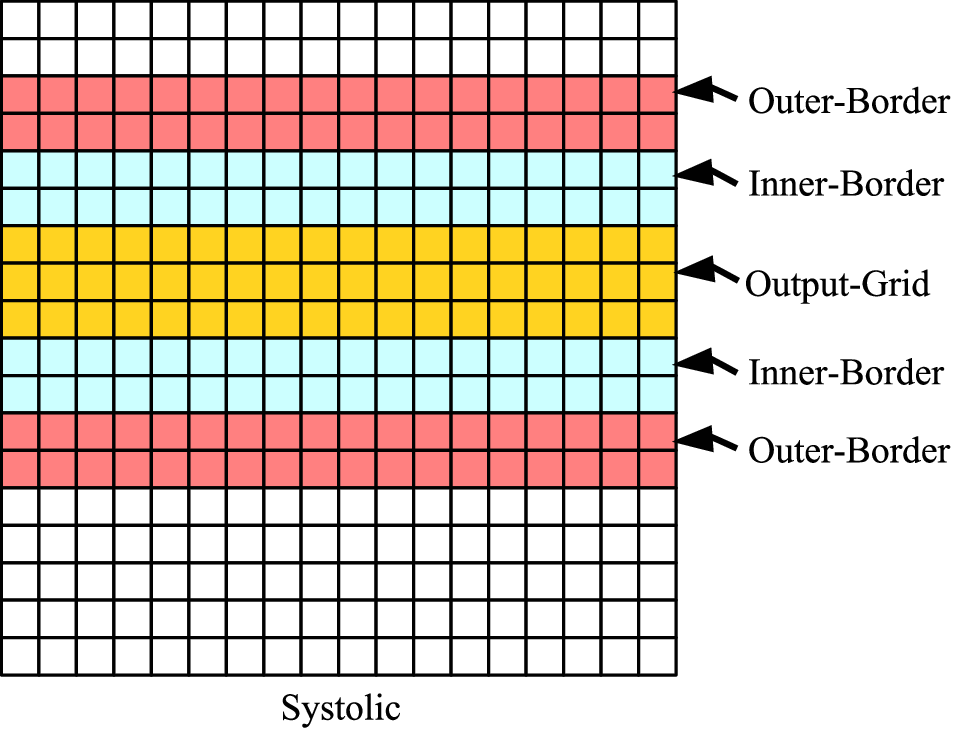,height=2.2in} \\
\end{tabular}
\caption{\label{fig:partial} The input sub-grid required to calculate DBSCAN of an~$I_R \times I_C$ sub-grid of a larger grid of events. In both pictures,~$I_R=3$, and $\epsilon=2$.  In the
flat picture,~$I_C$ is 6.  In the systolic picture, because systolic networks work on entire
columns,~$I_C$ equals all of the columns.}
\end{center}
\end{figure}

We illustrate with the left side of Figure~\ref{fig:partial}, 
which shows a concrete example where~$I_R=3$, 
$I_C=6$ and $\epsilon=2$. To calculate the DBSCAN of the Output-Grid, one needs to construct
a DBSCAN network that contains the Outer-Ring, Inner-Ring and Output-Grid.  The width of both
rings is~$\epsilon$.  The only output neurons are the {\em Core} and {\em Border} neurons for the
Output-Grid.  The $B$ and $Border$ neurons for both rings are unnecessary, and may be deleted
along with all of their pre and post synapses.  The~$C$ and $Core$ neurons are unnecessary
for the Outer-Ring, and may also be deleted along with all of their pre and post synapses.  
The $C$ and $Core$ neurons are necessary for the Inner-Ring; however the $Core$ neurons for the
Inner-Ring should not be output neurons, because there are no corresponding $Border$ neurons
for them.

The systolic networks are similar, with an example in the right side of Figure~\ref{fig:partial}.
Since systolic networks work on all of the columns of a network,~$I_C = C$.  As with
the flat network, in the systolic networks, the~$B$ and~$Border$ neurons are unnecessary on
the borders, and the~$C$ and~$Core$ neurons are unnecessary on the Outer-Border.

In both networks, if the rings/borders go outside of the input grid (for example, if the 
Output-Grid starts in row zero and column zero), then any neurons that correspond to them may
be deleted.  Alternatively, the neurons may be retained, and no events be spiked in as inputs.
Clearly, the former solution is more efficient; however the latter solution may be attractive
as a single neural network may used for all Output-Grids of the same size.

\section{Practical Considerations}

In this section, we discuss some practical considerations with neuromorphic 
implementations of DBSCAN.

\subsection{Sizes and Times}

In Table~\ref{tab:sizes}, we compare the two algorithms in terms of parameters that matter
in spiking neural networks.  In this table, we provide general values, and values specific
to two grid sizes: $10 \times 10$ with~$\epsilon=2$, which is a small but nontrivial grid size, 
and $260 \times 346$, which is 
the field of vision of the DAVIS346 event-based camera by 
Inivation~\cite{i:19:346}. In this latter example, we set~$\epsilon=4$ for a more 
robust DBSCAN calculation.  For the specific grid sizes, we do not use the equations to
calculate neurons and synapses, but instead construct the networks with our software and
report the neurons and synapses of those networks. 

\begin{table}[ht]
\begin{center}
{\small
\begin{tabular}{|l|ccc|ccc|}
\hline
 & & Flat & & & Systolic & \\
 & General & $10 \times 10$ & $260 \times 346$ &
   General & $10 \times 10$ & $260 \times 346$ \\
\hline
Neurons & $5RC$ & 500 & 449,800 & $R*(4\epsilon+5)$ & 130 & 5,460 \\
Synapses & $(3+2N)RC$ & 4,172 & 14,626,040 & $R(2N+4\epsilon+3)$ & 550 & 46,700 \\
Timesteps to Solution & 5 & 5 & 5 & $C+2\epsilon+4$ & 18 & 358 \\
Timesteps to Reuse & 1 & 1 & 1 & $C+2\epsilon$ & 14 & 354 \\
Max Synapse Delay & 4 & 4 & 4 & 4 & 4 & 4 \\
Max Neuron Threshold & $minPts-1$ & $minPts-1$ & $minPts-1$ & $minPts-1$ & $minPts-1$ & $minPts-1$ \\
Max Fan-Out & $N+1$ & 26 & 82 & $N+2$ & 27 & 83 \\
Max Fan-In & $N-1$ & 24 & 80 & $N-1$ & 24 & 80 \\
\hline
\end{tabular}}
\caption{\label{tab:sizes} Comparison of the flat and systolic networks on parameters of
interest in the utilization of spiking neural networks. }
\end{center}
\end{table}

\subsection{Rows vs.~Columns}

In the systolic algorithm, the choice to spike in columns of inputs rather than rows is
completely arbitrary.  When~$R=C$, there is no practical difference.  If~$R \ne C$, then
the choice to spike in columns rather than rows has some practical implications:

\begin{itemize}
\item If~$R > C$, then spiking in columns results in larger networks than spiking in rows.
\item If~$R > C$, then spiking in columns results in smaller trains of input spikes and faster
      calculation of DBSCAN.
\item It may be easier from a software perspective to spike in columns or rows, depending on
      how events are received from the outside world.
\end{itemize}

In the flat algorithm, all inputs are spiked in at the same time and the network size is a
function of~$R \times C$, so there is no difference between rows and columns.

\subsection{Leak}

The requirement of leak is so that neurons whose thresholds are greater than one may be reused
from timestep to timestep.  The relevant neurons are the~$C$ neurons, 
whose thresholds are~$minPts-1$, and the output neurons ($Core$ and $Border$) whose thresholds
are two.  The leak requirement can be relaxed with the addition of extra neurons, synapses
and timesteps.  The technique is straightforward and demonstrated with respect to the~$Core$
neurons in the flat construction. 

\begin{figure}[ht]
\begin{center}
\begin{tabular}{ccc}
\epsfig{figure=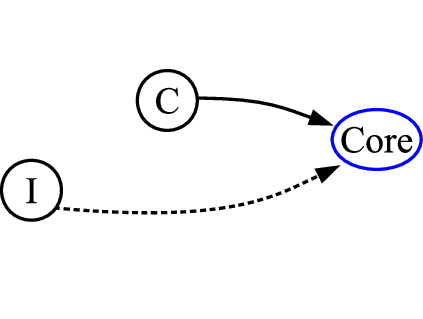,height=1.4in} &
\epsfig{figure=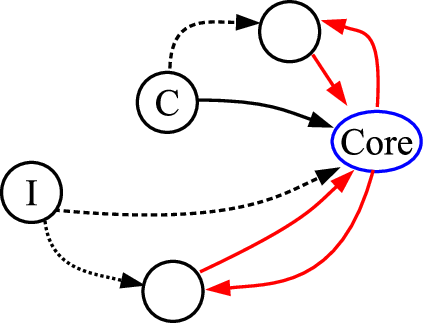,height=1.4in} &
\epsfig{figure=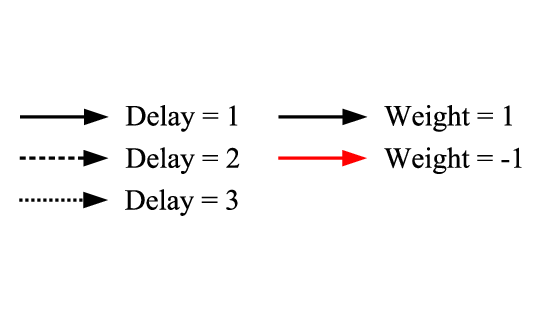,height=1.4in} \\
\end{tabular}
\caption{\label{fig:leak} A network with leak, converted to an equivalent network
without leak.  All neurons have thresholds of one, with the exception of the {\em Core}
neurons that have threshholds of two.}
\end{center}
\end{figure}

On the left side of Figure~\ref{fig:leak}, we show how {\em Core} neurons receive spikes from
their corresponding~$C$ and~$I$ neurons.  With leak, if the {\em Core} neuron does not fire, it
leaks its potential to zero at the end of a timestep.  Without leak, we need to add -1 charge
to the {\em Core} neuron if only one of~$C$ or~$I$ fire.  The network shown in the middle
of Figure~\ref{fig:leak} shows how to do that, although it adds extra timesteps to the
calculation and affects network reuse.
The reader is referred to~\cite{pzs:21:snn} for further examples that apply to these networks.

\subsection{Communication}

When rendered to a hardware neuroprocessor, communication is a very important consideration.
We give some concrete examples.  Suppose we are employing a neuroprocessor such as the
one described by Gupta, Vyas and Trivedi, implemented on a Xilinx Virtex 6 FPGA~\cite{gvt:20:fis}.
They report a capacity of 800 neurons and 12,544 synapses, executing at a speed of 100 MHz,
but with a time resolution of 2.5us.

Suppose we are processing a $10 \times 10$ grid, with~$\epsilon=2$ using the flat algorithm.  
From Table~\ref{tab:sizes}, we see that the network is 500 neurons and 4172 synapses, and
thus fits onto the neuroprocessor.  At 100 MHz, the neuroprocessor is capable of computing
DBSCAN in 50ns, processing a new grid every 10ns.  However, with a time resolution 
of 2.5us, communication of inputs and outputs will be the bottleneck.  Assuming a sparse
grid, with just 10 events, it will take 25us to get the input events into the neuroprocessor,
resulting in a speed of 40 KHz per DBSCAN calculation.

If we employ a systolic network, then the smaller network also fits onto the neuroprocessor.
A DBSCAN calculation requires 160ns to compute, which means that communication still the
bottleneck, and the speed of calculation remains 40 Khz.

If we consider the larger example, of a $260 \times 340$ grid, the flat algorithm is
completely intractable.  The systolic algorithm, on the other hand, requires 5,460 neurons
and 46,700 synapses, which is more than the limits of the neuroprocessor.
However, we can use the techniques described in Section~\ref{sec:partial} to
partition the problem into 10 sub-problems, working on Output-Grids of 26 rows.  
The resulting systolic network has 770 neurons and 5,554 synapses, and therefore fits
on the neuroprocessor.  Each DBSCAN calculation takes 358 timesteps, which is 3.58us per
calculation, or 35.8us for all ten networks.  The networks are identical, so they do not
require any reloading into the neuroprocessor.  Assuming the input grid contains more than
14 events, the communication of inputs will once again be the bottleneck.

\section{Open-Source Implementation}

Our implementation of these network constructions is available as open-source software
on~\url{https://github.com/TENNLab-UTK/dbscan}.  The networks produced may be executed
on the RISP neuroprocessor that is part of the TENNLab Open-Source Neuromorphic Computing
Framework (\url{https://github.com/TENNLab-UTK/framework-open} - slated for release in
October, 2024), with open-source FPGA implementation in~\url{https://github.com/TENNLab-UTK/fpga}).
The output format is simple enough for a straightforward translation to any LIF neuroprocessor.

\section{Conclusion}

We have detailed algorithms for neuromorphic implementations of the DBSCAN clustering algorithm.
This is an important algorithm from knowledge discovery and data mining that has applications
to the processing of event-based cameras, which motivates a neuromorphic implementation. 

There are two implementations -- a flat implementation 
that computes DBSCAN in 5 neuromorphic timesteps
but produces large networks, and a systolic implementation that takes longer to compute, but
produces much smaller networks.  We have specified these implementations precisely, analyzed them
in practical scenarios, and provide an open-source implementation.

Our future work includes loading these networks on the RISP neuroprocessor~\cite{pzg:22:risp},
both in microcontroller settings~\cite{pgf:22:dns} and on FPGA (\url{https://github.com/TENNLab-UTK/fpga}) to assess their effectiveness in real-time embedded settings, especially as front-end
filters for event-based cameras in neuromorphic control applications (e.g.~\cite{nlr:23:dotie,rsp:22:ebc}). There are likely other applications that will benefit from low-SWaP  spatial clustering and noise filtration, that don't involve event-based cameras, that can be explored. Although our spiking neural network implementations of the DBSCAN algorithm do not label individual clusters as the canonical DBSCAN algorithm does, we believe that, if desired, this functionality can be performed by a separate network downstream.

\bibliographystyle{plain}
\bibliography{bib}

\end{document}